%% file: cvpr.tex
\documentclass[final]{cvpr}

\usepackage{times}
\usepackage{epsfig}
\usepackage{graphicx}
\usepackage{amsmath}
\usepackage{amssymb}
\usepackage{xcolor}

\usepackage[pagebackref=true,breaklinks=true,colorlinks,bookmarks=false]{hyperref}

\def\cvprPaperID{13} 
\def\confYear{CVPR 2021}

\begin{document}

\title{Label Geometry Aware Discriminator for Conditional Generative Networks }

\author{Suman Sapkota\thanks{Equal contribution} \\ 
NAAMII, Nepal \\
\and 
Bidur Khanal\footnotemark[1] \\ 
NAAMII, Nepal \\
\and 
Binod Bhattarai\thanks{Corresponding author, b.bhattarai@imperial.ac.uk} \\
Imperial College London \\
\and 
Bishesh Khanal \\ 
NAAMII, Nepal \\ 
\and 
Tae-Kyun Kim\\
Imperial College London, UK and KAIST, South Korea\\
}


\maketitle

\input{abstract}
\input{intro}
\input{related_works}
\input{method}
\input{experiments}

\input{conclusion}

{\small
\bibliographystyle{ieee_fullname}
\bibliography{egbib}
}
\input{supplementary}
\end{document}

%% file: abstract.tex
\begin{abstract}
Multi-domain image-to-image translation with conditional Generative Adversarial Networks (GANs) can generate highly photo realistic images with desired target classes, yet these synthetic images have not always been helpful to improve downstream supervised tasks such as image classification.
Improving downstream tasks with synthetic examples requires generating images with high fidelity to the unknown conditional distribution of the target class, which many labeled conditional GANs attempt to achieve by adding soft-max cross-entropy loss based auxiliary classifier in the discriminator.
As recent studies suggest that the soft-max loss in Euclidean space of deep feature does not leverage their intrinsic angular distribution, we propose to replace this loss in auxiliary classifier with an additive angular margin (AAM) loss that takes benefit of the intrinsic angular distribution, and promotes intra-class compactness and inter-class separation to help generator synthesize high fidelity images.

We validate our method on RaFD and CIFAR-100, two challenging face expression and natural image classification data set.
Our method outperforms state-of-the-art methods in several different evaluation criteria including recently proposed GAN-train and GAN-test metrics designed to assess the impact of synthetic data on downstream classification task, assessing the usefulness in data augmentation for supervised tasks with prediction accuracy score and average confidence score, and the well known FID metric.
\end{abstract}

%% file: intro.tex
\section{Introduction}
\label{sec:intro}
Image-to-image translation task with conditional Generative Adversarial Networks (cGANs) has seen tremendous progress in recent years, synthesizing realistic images from input source images conditioned with target styles or domains~\cite{isola2017pix2pix,zhu2017unpaired,odena2017conditional,stargan_cvpr2018,he2019attgan,karras2019style}.
Such synthetic images are increasingly being used for augmentation in discriminative tasks for diverse applications including face recognition~\cite{gecer2018eccv}.

Although GANs can generate highly realistic-looking images these days, the high perceptual quality can be deceiving when these images are used for downstream tasks.
\emph{Seeing is not believing}~\cite{ravuri2019seeing} and ~\cite{evalgan_eccv2018} show that using synthetic images for downstream tasks do not necessarily improve their performance even when the quality of the synthetic images seem to match real image distribution qualitatively or under commonly used evaluation metrics like Inception score.
In order to better assess the quality of generated images for downstream tasks, ~\cite{evalgan_eccv2018} propose GAN-train and GAN-test metrics.
Some work try to improve the downstream tasks by exploring methods that selectively sample only a subset of generated images~\cite{bhattarai2020sampling,xue2020selective}. 
However, a separate sampler adds new set of hyperparameters or computation, and may be less attractive compared to generating better samples in the first place.
Generating better samples that can be used for augmentation of downstream task would require greater control over the target class attributes and achieve two important goals: \textit{fidelity} and \textit{coverage}.
In other words, synthetic examples for data augmentation should have high fidelity to the target class while preserving the remaining attributes, and be diverse enough following the real conditional data distribution over its full support.
This is particularly important in applications such as human face expressions where minor but very specific variation in images results in the image attribute to change to a specific class, e.g. contempt to anger. 

Several state-of-the-art labeled conditional GANs for multi-domain image-to-image translation use auxiliary soft-max classifiers~\cite{odena2017conditional,stargan_cvpr2018} in addition to the standard adversarial and reconstruction losses.
Table~\ref{tab:ablation_lambda_c} shows the impact of using different weights for auxiliary soft-max classification loss in a popular labeled condtional GAN, StarGAN~\cite{stargan_cvpr2018}.
The result highlights the importance of auxiliary classification loss with the variation in its weight significantly affecting the GAN performance.
We propose to improve this auxiliary classification component of labeled conditional GANs helping generator to synthesize better class conditional samples.
Inspired from recent progress in discriminative models for classification with the introduction of angular loss with margin~\cite{wang2018cosface,deng2019arcface,choi2020amc}, we replace the soft-max loss of the auxiliary classifier with the additive angular margin (AAM) loss~\cite{deng2019arcface}.
This modification helps the auxiliary classifier to learn representation that promote intra-class compactness and inter-class separation, and is better adapted to the intrinsic angular distribution of deep features~\cite{choi2020amc}.
Although various forms of angular loss and margins are possible, AAM loss has better geometric attribute with exact correspondence to the geodesic distance and perform better in supervised classification tasks~\cite{deng2019arcface}.
We hypothesize that these properties of the AAM loss are helpful in cGAN training as well, encouraging generator to improve the fidelity and coverage to the the class conditional distribution.
Moreover, as part of the AAM loss computation, we normalize the weights of the classifier to project it into a unit hypersphere which naturally lends itself to the stability of GAN training with bounded discriminator~\cite{gulrajani2017improved,miyato2018spectral,park2019sphere}.
Our results show that this simple but effective modification improves the performance of the state-of-the-art in several different metrics for evaluating GANs: 
GAN-Train, GAN-Test, FID, Average Confidence Score etc. (see Section~\ref{subsec:quanteval}).
Moreover, we show that augmenting downstream classification task using the images generated from the proposed method improves its robustness to noisy labels (Figure~\ref{fig:perf_label_distortion}).

Our contributions can be summarized as:\\
- to our knowledge, we are the first to propose additive angular loss with margin for conditional GAN setting, bringing the recent success of angular margin losses from supervised discriminator models to conditional GANs\\
- we improve state-of-the-art labeled conditional GANs in recently proposed metrics of GAN-test and GAN-train, particularly improving the performance of downstream tasks when using generated images for augmentation,\\
- show that learning representation that increases intra-class compactness and inter-class separation can improve robustness to label noise.

\begin{table*}[]
    \centering
    \begin{tabular}{l|l|l|l|l|l|l}
     Metric \textbackslash $\lambda_{cls}$  &  0.5 & 1.0 (default) & 1.5 & 2 & 5 & 10 \\
       \hline 
       GAN-train $\uparrow$ & 22.0 & 23.0 & 17.0 & 14.0 & 13.0 & 10.0 \\  
       \hline 
       GAN-test $\uparrow$ & 53.0 & 58.0 & 65.0 & 58.0 & 46.0 & 39.0 \\
       \hline
       FID $\downarrow$ & 58.0  &  66.7 & 108.9 & 110.3 & 117.4 & 133.9 \\  
       \hline
    \end{tabular}
    \caption{Performance of baseline method with varying weight of class cross-entropy loss on 
    Discriminator on CIFAR-100}
    \label{tab:ablation_lambda_c}
\end{table*}

%% file: related_works.tex
\section{Related Work}
\label{sec:related_works}
\paragraph{Multi-domain image-to-image translation with cGAN:}
Conditioning GANs has been an active research field ever since Mirza \textit{et al.}~\cite{mirza2014conditional} first showed that auxiliary information such as target label can be fed into standard GANs~\cite{goodfellow2014generative} to generate images conditioned on the auxiliary information.
Various forms of representations such as one-hot encoded target domain label~\cite{chen2016infogan,stargan_cvpr2018,he2019attgan,liu2019stgan,zhang2017age}, source domain in the form of semantic text or layout representations~\cite{reed2016generative,zhang2017stackgan,hong2018inferring,zhao2019image}, source information in the form of images~\cite{isola2017pix2pix,huang2018multimodal,liu2017unsupervised}, geometry captured in the form of sparse landmarks~\cite{zakharov2019few} have successfully been applied to control the nature of synthetic data.
Popular image-to-image translation works such as ~\cite{isola2017pix2pix,karras2019style} generate visually pleasing images, and may use reconstruction loss such as cycle consistency to preserve the major attributes of the source image~\cite{zhu2017unpaired}.
However, these methods still lack enough control over target class and do not have explicit constraint to generate images having high fidelity to target classes.

In order to improve the fidelity of the target attributes in synthetic images, in addition to conditioning the network with one-hot target vector, AC-GAN~\cite{odena2017conditional} first proposed adding an auxiliary classification loss to the discriminator.
Recent works have unified the ideas of standard adversarial loss, reconstruction loss and auxiliary classification loss to develop powerful class labeled image-to-image translation GANs with better control over target domains~\cite{he2019attgan,stargan_cvpr2018,liu2019stgan,bhattarai2020inducing,gecer2018eccv}.

\paragraph{Improving GAN Discriminator:}
The well known stability and convergence problems of GANs have led to several contributions towards better behaved GAN objective~\cite{arjovsky2017wasserstein}, regularization on weights and normalization~\cite{gulrajani2017improved,mescheder2018training,miyato2018spectral,park2019sphere}, consistency based regularization~\cite{chen2019self,zhang2020consistency}, or modifying the discriminator to learn more powerful representations~\cite{schonfeld2020u} and be robust against adversarial attacks~\cite{zhou2019don}.
The importance of discriminator is further reinforced by several recent contributions in conditional GAN settings as well.
Apart from ~\cite{miyato2018cgans}, who propose combining class vector information via inner product in discriminator instead of commonly used concatenation, most other recent work propose adding extra regularization loss or modifying the discriminator task including auxiliary classifiers.
TAC-GAN~\cite{gong2019twin} propose an extra classifier to add negative class conditional entropy term to AC-GAN to improve the density of generated samples near decision boundary of the classifier, while \cite{han2020unbiased} propose using mutual information between generated data and labels instead of adding the classifier.

\paragraph{Rethinking soft-max classification cross-entropy loss:}
The standard soft-max cross-entropy loss for auxiliary classifier do not explicitly take into account the relationship between data samples.
Center loss~\cite{wen2016discriminative} imposes the intra-class compactness but does not increase the inter-class separation.
Contrastive loss~\cite{sun2014deep} brings the similar points together and pushes the dissimilar ones, but it requires the paired or triplet examples.
Angular loss with margin has shown state-of-the-art results in supervised discriminative models while promoting intra-class compactness and inter-class separation, and leveraging the intrinsic angular distribution of deep features~\cite{wang2018cosface,deng2019arcface,choi2020amc}.
We build on this development and replace soft-max with additive angular margin loss (AAM)~\cite{deng2019arcface} in the auxiliary classifier of conditional GAN framework.
We show that increasing intra-class compactness and inter-class separation by using AAM loss helps generator to synthesize images with high class fidelity and coverage, and improving the performance of downstream tasks when used for augmentation. 
Concurrent works reinforce the potential of this direction to learn representation leveraging inductive bias about the relationship of data samples: \textit{the Rumi framework} in ~\cite{asokan2020teaching} learns a discriminator to classify images into positive-class real, negative-class-real and fake images, and ~\cite{kang2020contragan} use conditional contrastive loss to increase intra-class compactness and inter-class separation.
Our method is simple but effective that can be adapted to any auxiliary classifier based GANs.

%% file: method.tex
\section{Method}
\label{sec:method}
This section introduces class labeled conditional GAN followed by our proposed modification in discriminator with additive angular margin (AAM) loss.

\begin{figure*}[t]
\includegraphics[trim=1cm 5cm 1.7cm 1cm, clip, width=0.95\textwidth]{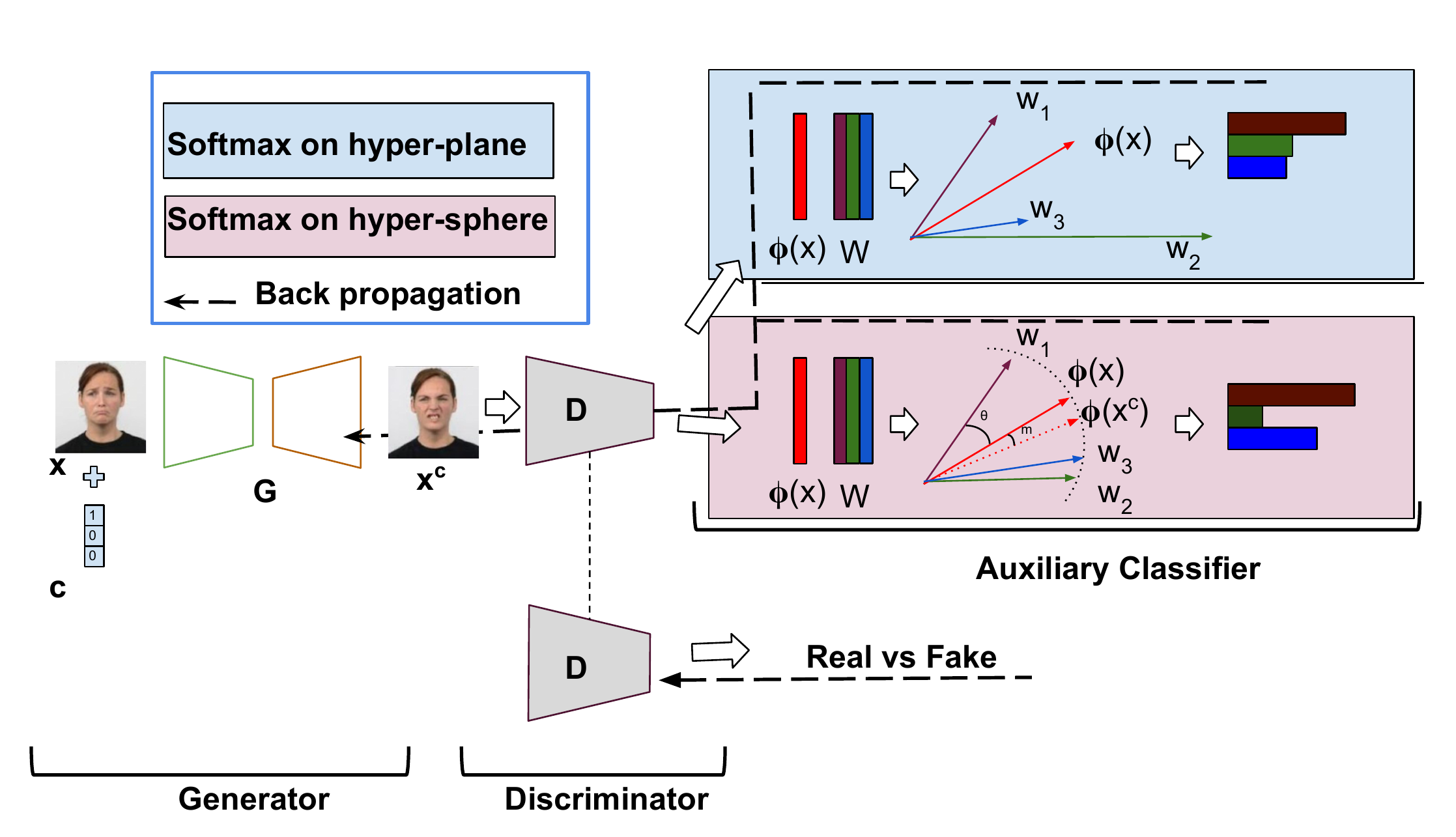} 
\caption {Overall pipeline of Modified Discriminator Classification Loss. We modified the original Softmax loss (right-top) in Auxiliary classifier by introducing  an additive angular margin (m) on its hyper-sphere projection (right-bottom).}
\label{fig:main_method}
\end{figure*}

\subsection{Class labeled conditional GAN}
In class labeled conditional GAN, given an input image $x$ and target category information $c$, the objective is to translate $x$ faithfully to a new image $x^c$ of the target category.
We consider the case where $c$ is prescribed using one-hot vector representation, the discriminator D is equipped with a separate branch for auxiliary classifiers, and a single generator G is trained to generate images of multiple categories, i.e. $x^c = \text{G}(x, c)$ (see Figure~\ref{fig:main_method}).

Let us consider a labeled training dataset, $\mathcal{D} = \{(x_i, y_i)\, |\, i = 1, 2, ..., N\}$, where $x_i \in \mathbb{R}^{W \times H \times 3}$ is an input image of width $W$ and height $H$ of category $y_i \in \mathbb{R}^{C}$ represented as one-hot encoded vector.
$C$ and $N$ are the number of categories and total images in the training set respectively.
Discriminator D estimates probability distribution over images with $D_{src}(x)$ giving probability that $x$ is real, and over category labels with $D_{cls}(y|x)$ giving probability that $x$ belongs to category $y$.
The goal is to use this D in GAN framework such that the images generated by generator G, $x^c = G(x,c)$ follows the unknown true distribution $P(x^c|c)$ while preserving remaining attributes of $x$.
Training the network depends on three loss components described below.

\noindent \textbf{Adversarial Loss}
This is similar to the standard unsupervised loss where D learns to identify real vs fake images, and G learns to fool D by producing realistic images.
As this is a conditional image-to-image translation GAN, one difference from general GANs is that the generator takes as input a source image $x$ and target class $c$.
\begin{equation}
\begin{split}
 \mathcal{L}_{adv} = & \thinspace {\mathbb{E}}_{x} \left[ \log{{D}_{src}(x)} \right]  \> \>  +   \\
 & \thinspace {\mathbb{E}}_{x, c}[\log{(1 - {D}_{src}(G(x, c)))}],
\end{split}
\label{adversarial_loss}
\end{equation}

\noindent \textbf{Reconstruction Loss}
In order to preserve the content of the source image $x$ that are not relevant or invariant to target class $c$, we add a cycle consistency loss where G is encouraged to successfully synthesize back $x$ when $x^c = G(x,c)$ is fed as source image to G together with the original label $y$.
\begin{equation}
\mathcal{L}_{rec} = {\mathbb{E}}_{x, c, y} [{||x - G(G(x, c), y)||}_{1} ],
\label{eqn:recon}
\end{equation}
\noindent \textbf{Discriminator Classifier Loss}
In order to encourage preservation of target category, two losses corresponding to real images and fake images are incorporated.
When using the real labeled data $(x,y)$, D is trained to minimize the classification loss
\begin{equation}
\mathcal{L}_{cls}^{r} = {\mathbb{E}}_{x, y}[-\log{{D}_{cls}(y|x)}],
\label{eqn:cls-real}
\end{equation}
\
and we use synthetic or fake images to train G with
\begin{equation}
\mathcal{L}_{cls}^{f} ={\mathbb{E}}_{x, c}[-\log{{D}_{cls}(c|G(x, c))}].
\label{eqn:cls-fake}
\end{equation}
The classification loss $\log D_{cls}(y|x)$ and $\log{{D}_{cls}(c|G(x, c))}$ in previous works use soft-max cross entropy loss, which we propose to replace with a new angular loss with margin (see Section~\ref{subsec:aam-loss}).

\noindent \textbf{Combined Loss}
The three losses are combined to train conditional GAN in a minmax game where we train D by minimizing
\begin{equation}
\mathcal{L}_{D} =  - \mathcal {L}_{adv} +  {\lambda}_{cls}\thinspace\mathcal{L}_{cls}^{r}
\label{eqn:discriminator_obj},
\end{equation}
and train G by maximizing
\begin{equation}
\mathcal{L}_{G} =   \mathcal {L}_{adv} +  {\lambda}_{cls}\thinspace\mathcal{L}_{cls}^{f} + 
{\lambda}_{rec}\thinspace\mathcal{L}_{rec}.
\label{eqn:generator_obj}
\end{equation}

While several works improved training stability and realism  of synthetic images by modifying Discriminator with regards to the adversarial loss component~\cite{arjovsky2017wasserstein,gulrajani2017improved,miyato2018spectral,park2019sphere}, few concurrent works have started exploring to improve conditional GAN by modifying auxiliary loss component~\cite{gong2019twin,han2020unbiased,kang2020contragan}.
We focus on improving this aspect of the GAN training.

\subsection{Angular Attributes Classification Loss}
\label{subsec:aam-loss}
As seen in Table~\ref{tab:ablation_lambda_c}, the auxiliary classification loss plays an important role in generating higher quality samples.
The gradients from the auxiliary loss are used not just in training D but propagate to G as well when maximizing G's objective in Equation~\ref{eqn:generator_obj} (also see Figure~\ref{fig:main_method}).
Most of the existing labeled conditional GAN including AC-GAN, StarGAN, STGAN, AttGAN use the soft-max classifier with cross-entropy loss function for $D_{cls}$ in Equations~\ref{eqn:cls-real} and~\ref{eqn:cls-fake} given by,
\begin{equation}
L_{s}= - \frac{1}{N} \sum_{i=1}^{N} \log \frac{e^{W^{T}_{y_{i}}\phi({x_{i}}) + b_{y_{i}}}}{\sum_{j=1}^{C} e^{W^{T}_{j}\phi(x_{i}) + b_{j}}} 
\label{eqn:softmax_hyperplane}
\end{equation}
where $\phi(x_{i}) \in \mathbb{R} ^{d}$ denotes the deep feature of the input sample $x_i$ having class label $y_i$.
$W \in \mathbb{R} ^{d\times C}$ is the weight matrix where $C$ is the number of classes, $W_{j} \in \mathbb{R} ^{d}$ the weight vector in column $j$ and $b_{j} \in \mathbb{R} ^n$ the bias vector.
We propose to make the auxiliary classifier more powerful by taking advantage of the intrinsic angular distribution of the deep features, and to learn the representation that increases both the intra-class compactness and inter-class separation.
We set the bias vectors $b_j$ to $0$, normalize the weight vectors $W_j$ to unit magnitude, and normalize deep feature vector $\phi(x_i)$ to unit magnitude and scale it by $s$.
Thus, the feature embedding are distributed in a hypersphere of radius $s$ and the output predictions are dependent only on the angle between the weight vectors and feature vector given by $\theta_j = \arccos\left(W^{T}_{j}\phi{(x_i)}\right)$.
As shown in Figure~\ref{fig:main_method}, this makes the embedding features to be distributed around a center of the surface area representing each category.
In order to improve the intra-class compactness and inter-class separation, we add an additional angular margin $m$ between $x_i$ and $W_{y_i}$.
This leads to the modification of the standard soft-max cross entropy loss in Equation~\ref{eqn:softmax_hyperplane} to additive angular margin (AAM) loss given by,
\begin{equation}
L_{ang}= - \frac{1}{N} \sum_{i=1}^{N} \log \frac{e^{{s}{(\cos(\theta_{y_{i}} + m ))}}}{{e^{{s}{(\cos(\theta_{y_{i}} + m ))}}}+\sum_{j=1, j \neq y_{i}}^{C}e^{s{\cos\theta_{j}}}} 
\label{eqn:angular_loss}
\end{equation}
We use this modified AAM loss in the auxiliary classification component of the discriminator $D_{cls}$ in Equations~\ref{eqn:cls-real} and ~\ref{eqn:cls-fake}.

%% file: experiments.tex
\section{Experiments}
\label{sec:experiments}
\subsection{Dataset}
\label{subsec:dataset}
We evaluate the proposed method on two different dataset: facial expression images,
\textbf{RaFD}~\cite{langner2010presentation}, and natural images with 100 classes,
\textbf{CIFAR-100}~\cite{krizhevsky2009learning}.
We chose these dataset with distinct characteristics to validate that we can benefit from the angular distribution of deep features of natural images in different settings, and that the proposed method is not limited to specific data domain.
\textbf{RaFD} consists of 8040 face images annotated with eight different facial expressions.
We split the dataset into two equal halves as training and test sets.
The images are aligned based on their landmarks and cropped to the size of $128 \times 128$.
~\textbf{CIFAR-100} is a challenging dataset for natural image classification problem consisting of 50K training and 10K test examples from 100 different classes.
Another reason to choose this dataset is that it has sufficiently large number of classes while not being too big such that the experiments can be run with reasonable computational resources.

\subsection{Implementation Details}
We implemented our algorithms on PyTorch. The experiments were carried out in a machine with an i7 processor and Titan 1070 Geforce GPU in Ubuntu Operating System. We train our model for 200K iterations. While training we set the initial learning rate as of 0.0001 for both Generator and Discriminator.
The learning rate is decayed after every 10K iterations.
Please check our code in the supplementary for the further 
implementation details.
We make our code public upon the acceptance of the paper.
\subsection{Compared Methods} We compared our method with several other class conditional GAN baselines from both the natural image domain and facial image domain.
Methods such as SNGAN~\cite{miyato2018spectral}, WGAN-GP~\cite{gulrajani2017improved}, DCGAN~\cite{radford2015unsupervised}, PixelCNN++~\cite{salimans2017pixelcnn++} are extensively evaluated on natural image domain. Similarly, 
StarGAN~\cite{stargan_cvpr2018}, EF-GAN~\cite{wu2020cascade}, Ganimation~\cite{pumarola2018ganimation} are mostly evaluated on face benchmarks.
We compared our method with these methods either quantitatively or qualitatively or both.
Among these baselines, StarGAN is the closest one as its architecture is designed to directly address the problem of target label retention on synthetic examples.
Other comparable architectures are AttGAN~\cite{he2019attgan}, STGAN~\cite{liu2019stgan}.
Recent extensive comparative study~\cite{bhattarai2020inducing} on the performance of StarGAN, AttGAN, STGAN on different conditioning methods identified similar trends on these architectures.
Thus, we expect similar impact of our method on AttGAN and STGAN as in StarGAN, and hence use StarGAN as the baseline for most of the comparisons in this work.

EF-GAN and Ganimation are not directly comparable.
EF-GAN trains multiple number of local GANs along with a global GAN in an hierarchical fashion. 
Whereas, Ganimation uses semantic masks as an extra information.
Our method does not need any extra parameters.
Other indirectly comparable methods are SNGAN, WGAN-GP as they do not directly address the problem of preserving labels on synthetic data directly, and focus on bringing stability in training with regularization in discriminator weights.
Thus, these are complementary to our methods and their impact on the proposed method may be assessed independently.
While this is an interesting direction, it is beyond the scope of current work. 
TAC-GAN~\cite{gong2019twin} and Projection Discriminator~\cite{miyato2018cgans} tackle the problem similar to ours.
However, TAC-GAN demands extra learnable parameters equal to that of a discriminator. 
Projection Discriminator~\cite{miyato2018cgans} injects label information to intermediate layers and learn the parameters in adversarial fashion without using an auxiliary loss.
Our focus is on improving the auxiliary classifier component of the GAN.
Even then, we have compared with some of the representative works from this category see where our method stands in terms of performance with contemporary approaches.

\begin{table}[]
    \centering
    \begin{tabular}{l|c|c}
        Method & GAN-Train$\uparrow$ & GAN-Test$\uparrow$ \\
        \hline 
         SNGAN~\cite{miyato2018spectral} & \textcolor{red}{45.0} & \textcolor{blue}{59.4} \\ 
         WGAN-GP(10M)~\cite{gulrajani2017improved} & 26.7 & 40.4\\ 
         WGAN-GP(2.5M)~\cite{gulrajani2017improved} & 5.4 & 4.3 \\ 
         DCGAN~\cite{radford2015unsupervised} & 3.5 & 2.4 \\ 
         PixelCNN++~\cite{salimans2017pixelcnn++} & 4.8 & 27.5 \\
         \hline 
         StarGAN~\cite{stargan_cvpr2018} & 23.0 & 58.3 \\
         Ours & \textcolor{blue}{30.0 (+7.0)} & \textcolor{red}{71.8 (+13.5)} \\
         \hline 
    \end{tabular}
    \caption{Comparison of GAN-Train and GAN-Test on CIFAR-100. We compared our method
    with some of the recent GAN architectures extensively evaluated on natural image generation.
    We use average image class recognition rate to measure the performance. Our method obtains highest and the second-highest performance on GAN-Test and GAN-Train respectively.
}
    \label{tab:gan_tt_cifar}
\end{table}

\begin{table}[]
    \centering
    \begin{tabular}{c|c|c}
        Method & GAN-Train$\uparrow$ & GAN-Test$\uparrow$ \\
        \hline 
        Ganimation~\cite{pumarola2018ganimation} & 84.36 & \textbackslash \\
        EF-GAN~\cite{wu2020cascade} & 89.38 & \textbackslash \\ 
        \hline 
         StarGAN~\cite{stargan_cvpr2018} & 91.33 & 89.85 \\
         Ours & \textcolor{red}{93.2 (+1.87)} & \textcolor{red}{95.16 (+5.31)} \\
         \hline 
    \end{tabular}
    \caption{Comparison of GAN-Train and GAN-Test on RaFD data set. We use average attribute recognition 
    rate to compare the performance of our method
    with three state-of-the-art methods for expressions translation. Our results show highest performance on both GAN-Train (vs.all three) and GAN-Test (vs. StarGAN).}
    \label{tab:gan_tt_rafd}
\end{table}

\begin{figure*}
    \centering
    \includegraphics[trim=0cm 0cm 0cm 0cm, clip, width=0.90\textwidth]{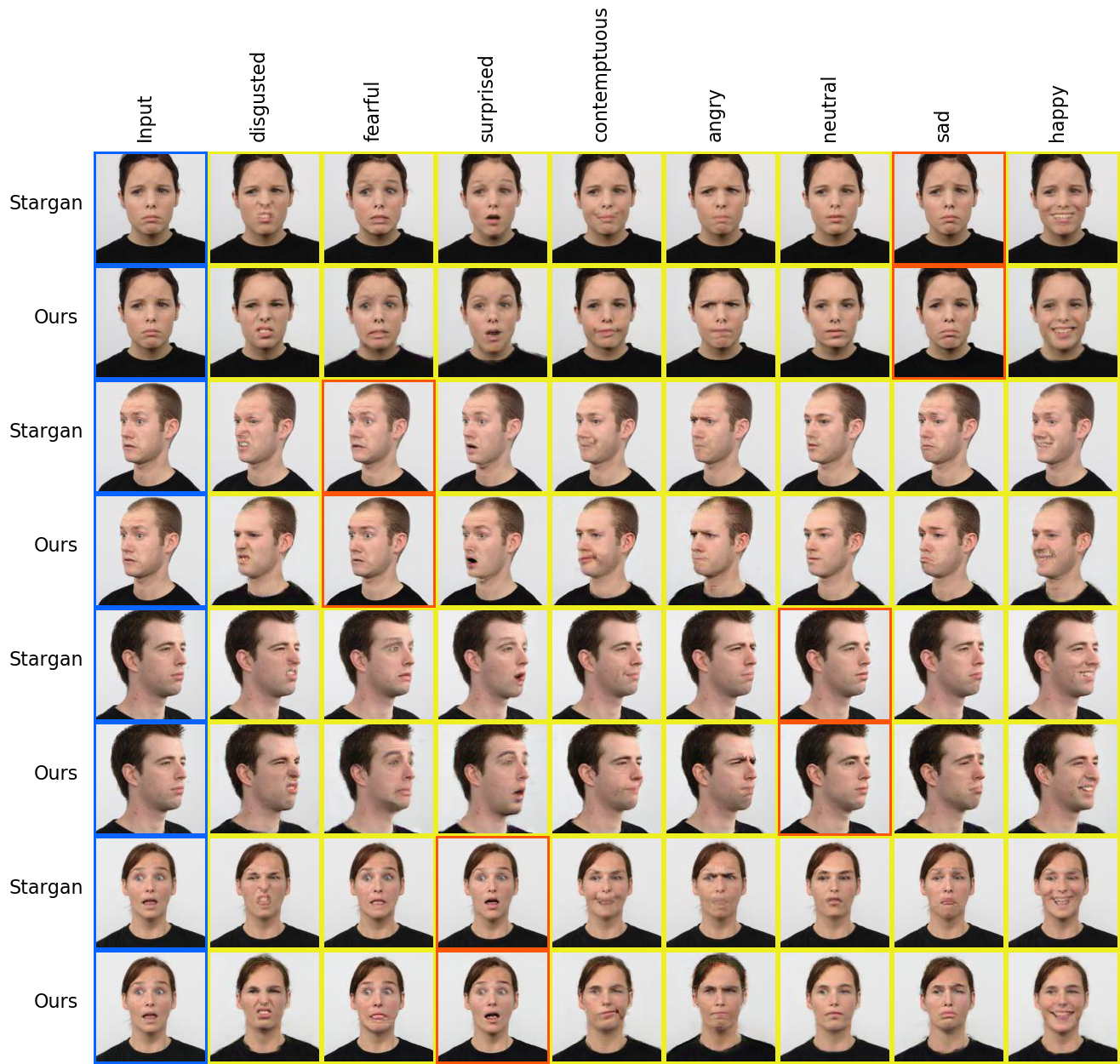}
    \caption{Qualitative comparison of face expression translation on RaFD dataset. Here, 
    first column (blue) shows the input image, red border shows the reconstruction and the rest of the column (yellow) shows its translated images.
    Label in each column starting from second column encodes the target attribute. Label on
    the rows are the applied methods. From these qualitative comparison, we see less artifacts and better preservation of the target attributes.}
    \label{fig:qual_rafd2}
\end{figure*}

\begin{figure}
    \centering
    \includegraphics[width=0.4990\textwidth]{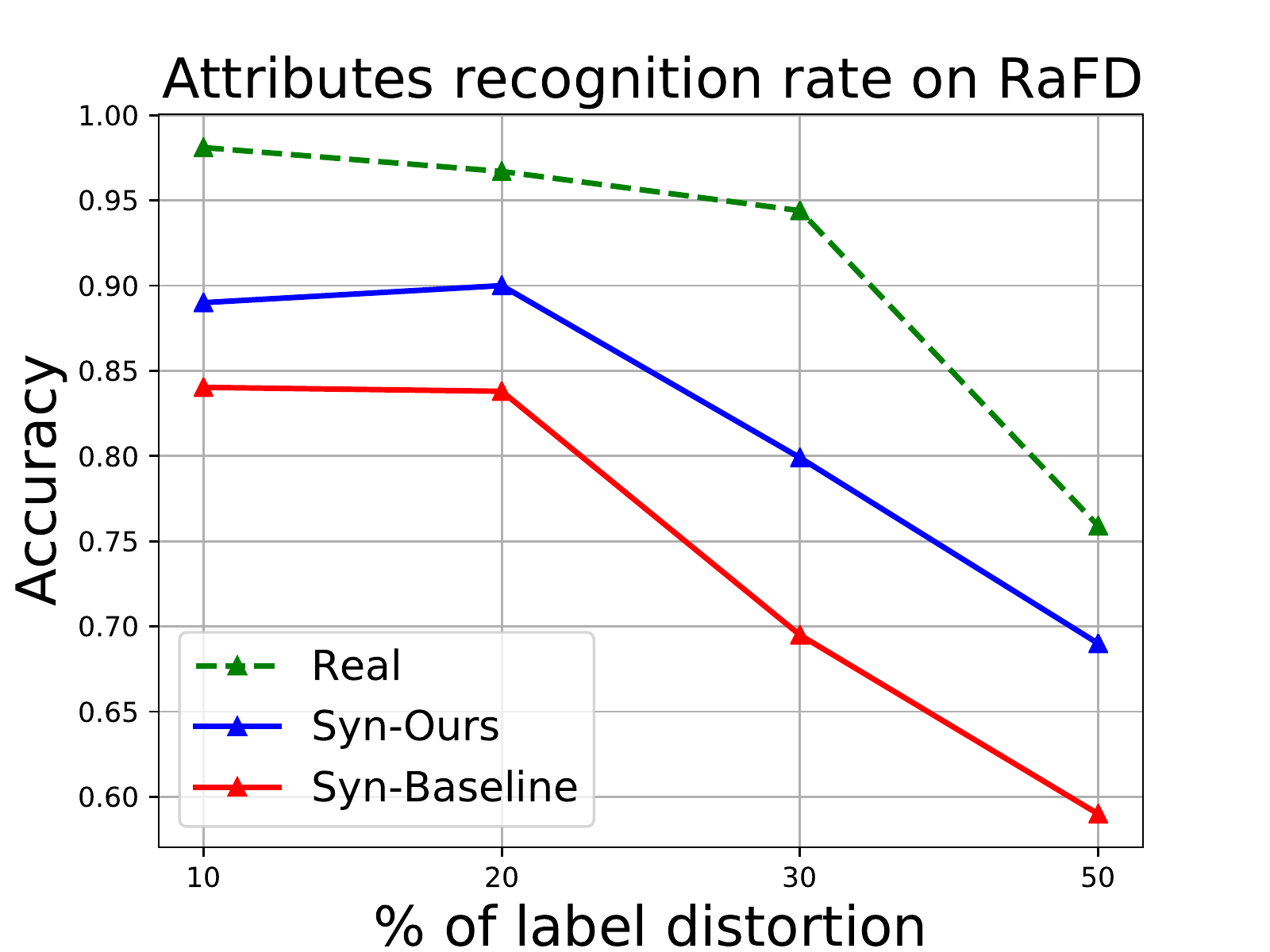}
    \caption{Comparison of attribute recognition rate on GAN-synthetic data 
    by the model trained with different proportion of labels distorted. 
    The distortion on label makes the classifier weak. This
    evaluates how robust are the labels on the synthetic data. Syn-ours
    is the performance of our method. Whereas, Syn-Baseline is the performance of StarGAN.
    }
    \label{fig:perf_label_distortion}
\end{figure}

\subsection{Quantitative Evaluation} 
\label{subsec:quanteval}
\noindent \textbf{GAN-Train and GAN-Test:} 
These are recently proposed metrics to evaluate the performance of GANs~\cite{evalgan_eccv2018}.
In GAN-train, a classifier is trained using synthetic images as input and the target labels used to generate corresponding images as ground truth label.
GAN-train is this classifier's prediction score on real test set.
Similarly, GAN-test is the prediction score of a classifier on a synthetic test set when the classifier is trained using a real images.
This is equivalent to attribute generation rate calculated
on the recent labeled conditional GANs~\cite{liu2019stgan,bhattarai2020inducing}.

To evaluate GAN-Test on CIFAR-100, we train ResNet-18~\cite{resnet_cvpr2016} on 
CIFAR-100 train set. The performance of this model on CIFAR-100 test set is 77\%.
Table~\ref{tab:gan_tt_cifar} 
shows the performance comparison of the proposed method with several directly and indirectly comparable contemporary methods for CIFAR-100.
As we see in the table, our method makes a substantial improvement over the baseline by +7\% on GAN-train and +13.5\% on GAN-test.
Also, our method out-performs almost every compared method in both the metrics, next to SNGAN in GAN-Train.
As mentioned before, SNGAN focuses in improving discriminator by regularizing every layer without a particular focus on auxiliary classification for class retention problem. 
Combining such regularization approach of SNGAN with our method may further improve the performance, and is left as future work. 
Performance of most of the methods are in lower band due to highly challenging in nature of CIFAR-100. 
Since the GAN-train and GAN-test are akin to two competing metrics of precision and recall, consistent performance on both the metrics validates that the proposed method is effective in handling a data set with large number of classes. 

Similarly, Table~\ref{tab:gan_tt_rafd} shows the performance comparison on RaFD.
Our approach surpasses the performance of the StarGAN by +1.87\% on GAN-train and +13.5\% on GAN-test.
In addition to this, our method also outperforms Ganimation and 
EF-GAN which are state-of-the-art methods in performing conditional GANs for expressions translation.
This consistency in performance across two contrasting nature of benchmarks clearly validates the benefit of the proposed method.
It also shows that the proposed angular geometric constraint is better at capturing the underlying distribution of data independent to the choice of particular image dataset benchmark.\\
\noindent\textbf{Performance on weak classifier:} We compare GAN-test score further using a weak classifier which is trained on real data with different proportion of labels distorted. Figure~\ref{fig:perf_label_distortion} shows that the performance on classifiers trained with noisy label is better when using proposed method compared to the baseline of StarGAN.

\begin{figure*}
    \centering
    \includegraphics[width=0.95\textwidth]{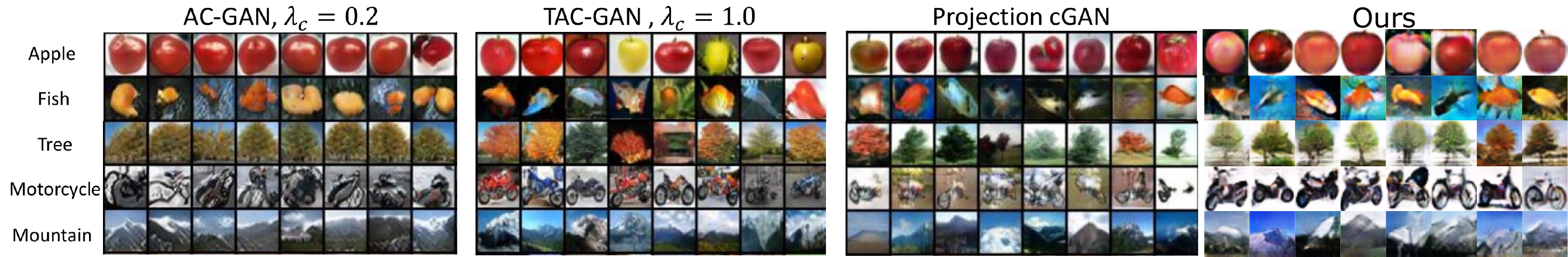}
    \caption{Here we make qualitative comparison of some of the recent methods with our proposed method. These are the randomly sampled images from five different categories. From these synthetic examples, we can clearly see our method generating images with lesser noise, better contrast and diversities too. If we carefully observe the shape of the apples, we can see the shape of the apples are distorted by all the existing methods. Whereas, our method managed to preserved its shape well. }
    \label{fig:qual_compare_cifar100}
\end{figure*}

\begin{table}[]
    \centering
    \begin{tabular}{c|c|c}
       Dataset & Method & FID (Backbone) $\downarrow$ \\
        \hline 
        RaFD & Ganimation~\cite{pumarola2018ganimation} & 45.55 (Inception V4) \\
        RaFD &  EF-GAN~\cite{wu2020cascade} & 42.36 (Inception V4) \\ 
        \hline 
        RaFD &  StarGAN~\cite{stargan_cvpr2018} & 6.14 (Inception V3)\\
        RaFD & Ours & 5.35 (Inception V3) \\
        RaFD &  StarGAN~\cite{stargan_cvpr2018} & 7.89 (Inception V4) \\
        RaFD & Ours & 6.17 (Inception V4) \\
        \hline
        CIFAR-100 &  StarGAN~\cite{stargan_cvpr2018} & 66.74 (Inception V3)\\
        CIFAR-100 & Ours & 53.73 (Inception V3) \\
        \hline 
    \end{tabular}
    \caption{FID comparison on both CIFAR-100 and RaFD. We computed the FID taking both 
    InceptionV3 and InceptionV4 as backbone. Our method outperforms these competitive baselines.
    }
    \label{tab:fid_comparison}
\end{table}


\vspace{-0.2cm}
\paragraph{Fr\'echet Inception Distance (FID):} FID ~\cite{unterthiner2017coulomb} is a widely used metric that measures the similarity of synthetic image statistics to the real image statistics using features of real and fake images.
Lower FID values correspond to better quality of fake images of  GAN evaluation that measures  
Table~\ref{tab:fid_comparison} shows the FID comparison on both RAFD
and CIFAR-100.
On both the benchmarks our method improves performance over the baseline which shows that imposing angular distribution geometry constraint in order to faithfully retain the target label on translated images does not hurt the other quality metric.

\begin{figure}
    \centering
    \includegraphics[width=0.5\textwidth]{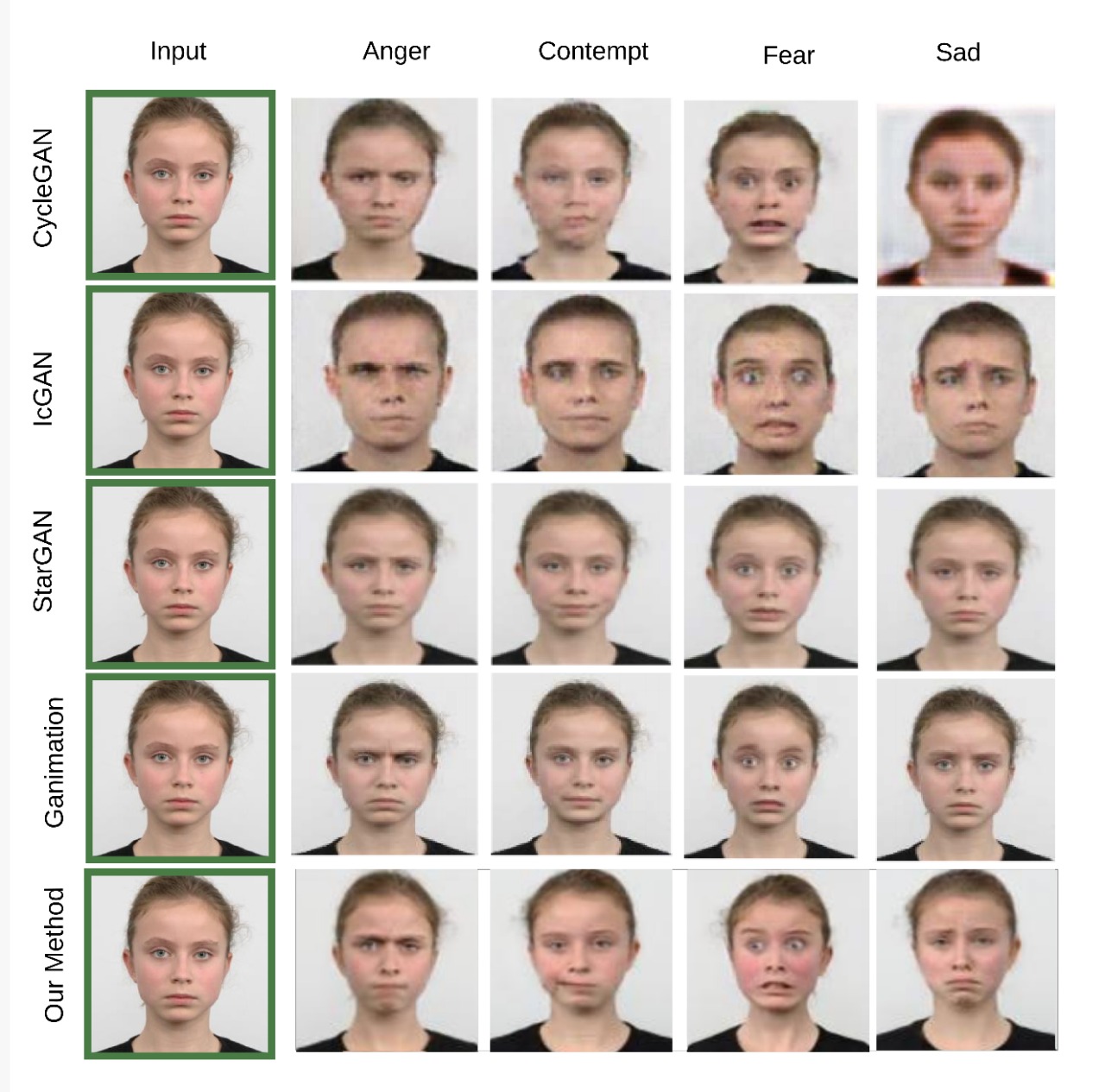}
    \caption{Qualitative comparison of attribute translation on RaFD. In comparison to the previous methods, our translations are more intense with less artifacts and better contrast.}
    \label{fig:qual_rafd1}
\end{figure}

\begin{figure}
    \centering
    \includegraphics[width=0.23\textwidth]{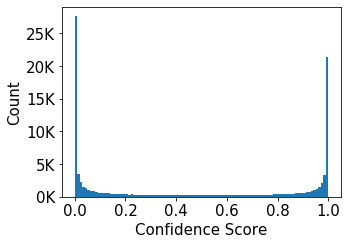}
    \includegraphics[width=0.23\textwidth]{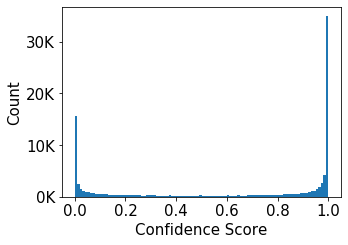}
    \caption{Distribution of confidence scores on  GAN synthetic image on CIFAR-100 : 
    StarGAN (Left) and Ours (Right).}
    \label{fig:conf_distribution}
\end{figure}

\begin{table*}[]
    \centering
    \begin{tabular}{l|l|l|l|l|l|l}
    \hline 
     Setup \textbackslash Real data size(K) & \textbf{2.5} & \textbf{5.0} & \textbf{10.0} \\
       \hline 
      Real only$!$~\cite{evalgan_eccv2018}  & 25.6 & 40.0 & 51.5  \\
      \hline
      Real only  & 24.39 & 36.32 & \textcolor{blue}{51.13}  \\  
       \hline 
       Real + Syn (Stargan, 50K) & \textcolor{blue}{31.45 (+7.06)} & \textcolor{blue}{39.65 (+3.3)} & 49.22 (-1.91) \\
       \hline
       Real + Syn (\textbf{Ours}, 50K) & \textcolor{red}{37.25 (+12.86)} & \textcolor{red}{44.20 (+7.98)} & \textcolor{red}{51.83 (+0.7)}  \\
       \hline
    \end{tabular}
    \caption{Performance comparison of data augmentation on CIFAR-100 using  using average attribute classification accuracy (higher is better).
    We followed the experiment setup proposed in~\cite{evalgan_eccv2018}. Real only! are the numbers reported by them, whereas Real only are from our implementation. We augmented 50K synthetic data with the variable size of real training examples shown in the top row.}
    \label{tab:data_augmantation}
\end{table*}

\paragraph{Data Augmentation:} 
This approach evaluates the benefit of adding synthetic images as part of data augmentation when training a classifier model with a real training data.
The evaluation metric measures the attribute classification performance of a model on a real test set, where the model is trained using real data and data augmentation with synthetic data.

Table~\ref{tab:data_augmantation} reports the performance comparison on data augmentation using average attribute classification accuracy on 
real test examples of CIFAR-100.
We followed the experimental setup proposed in~\cite{shmelkov2018good} 
to evaluate the performance.
\textit{Real only!} are the numbers reported in~\cite{shmelkov2018good} and \textit{Real only} are the results obtained in our implementation which are slightly lower but still comparable.
We augmented the variable size of real training examples 2.5K, 5K and 10K with 50K of synthetic data obtained from StarGAN and our method.
The table reports the performance of a classification model trained on each of these combinations of real and synthetic data.
We can see that the classifier trained on the data augmented with our synthetic data clearly outperforms the model trained on the data augmented with StarGAN.
The interesting point to note here is, when the real data set size is 10K,
augmenting StarGAN synthetic data with real data gives lower performance in comparison using only real data.
The performance using our synthetic example is still higher than using only real images, providing evidence that our method is able to translate more faithfully the images with better preserved target label.
However, it is still an open problem to improve by large margin the performance of discriminative model using data augmentation from synthetic example when very large number of real examples are available.

\paragraph{Average confidence score:} 
This is another recently proposed metric ~\cite{liu2019conditional} as 
attribute preserving for conditional GAN.
Here, class conditional probability on GAN-test examples are 
evaluated by a model trained on real training examples.
We trained a preactivation variant ResNet-18 ~\cite{he2016identity} on CIFAR-100 (just mentioned before)
and applied on 100K GAN-synthetic data. Figure~\ref{fig:conf_distribution}
compares the confidence score distribution of GAN synthetic test examples.
Ideally, all the mass should concentrate on right side (confidence 1) and 
make a single pole. In practice, our method has higher mass on right side than 
the StarGAN. This demonstrates that our method generates images more faithfully 
and confidently. It can be argued that generating an easy example may improve such metric.
However, performance of our method on other metrics such as GAN-train, FID
easily refutes it.

\paragraph{Qualitative Evaluations} For qualitative evaluations on CIFAR-100, 
please refer Figure~\ref{fig:qual_compare_cifar100} with inline explanation on caption.
Similarly, we present qualitative comparisons with Ganimation, StarGAN, CycleGAN 
on RaFD in Figure~\ref{fig:qual_rafd1}. Similarly, find additional qualitative comparisons 
on RaFD in Figure~\ref{fig:qual_rafd2}.

%% file: conclusion.tex
\section{Conclusion}
The proposed AAM loss in auxiliary classifier of labeled conditional GANs with its nice geometrical properties help synthesize high fidelity target class which is an important problem in face data applications.
Some recent methods like TAC-GAN~\cite{gong2019twin} and UAC-GAN~\cite{han2020unbiased} use extra constraint to prevent generator generate only ``easy" samples for the auxiliary classifer away from its decision boundary.
We believe that the issue may be less severe with our propsoed method where we make the classifier stronger and map the real images into a manifold with high intra-class compactness and large margin to separate inter-class samples.
Several experiments and evaluation metrics show that this geometry aware loss leading to better representation helps generator synthesize higher quality images preserving better the target labels.
While better representation learning with inductive bias on the manifold of conditional data distribution looks promising, it is still an open problem to completely understand the dynamics of conditional GAN training, where generator and discriminators equipped with classifiers compete in the non-stationary environment, and have complex interaction between learned representation of each network component and the various regularization constraints added to the system for training stability.
Detailed analysis of all these components with ablation studies to understand their impact would be an interesting future direction to pursue.

%% file: supplementary.tex

\appendix

\begin{figure*}
    \centering
    \includegraphics[trim=0cm 0cm 0cm 0cm, clip, width=1.0\textwidth]{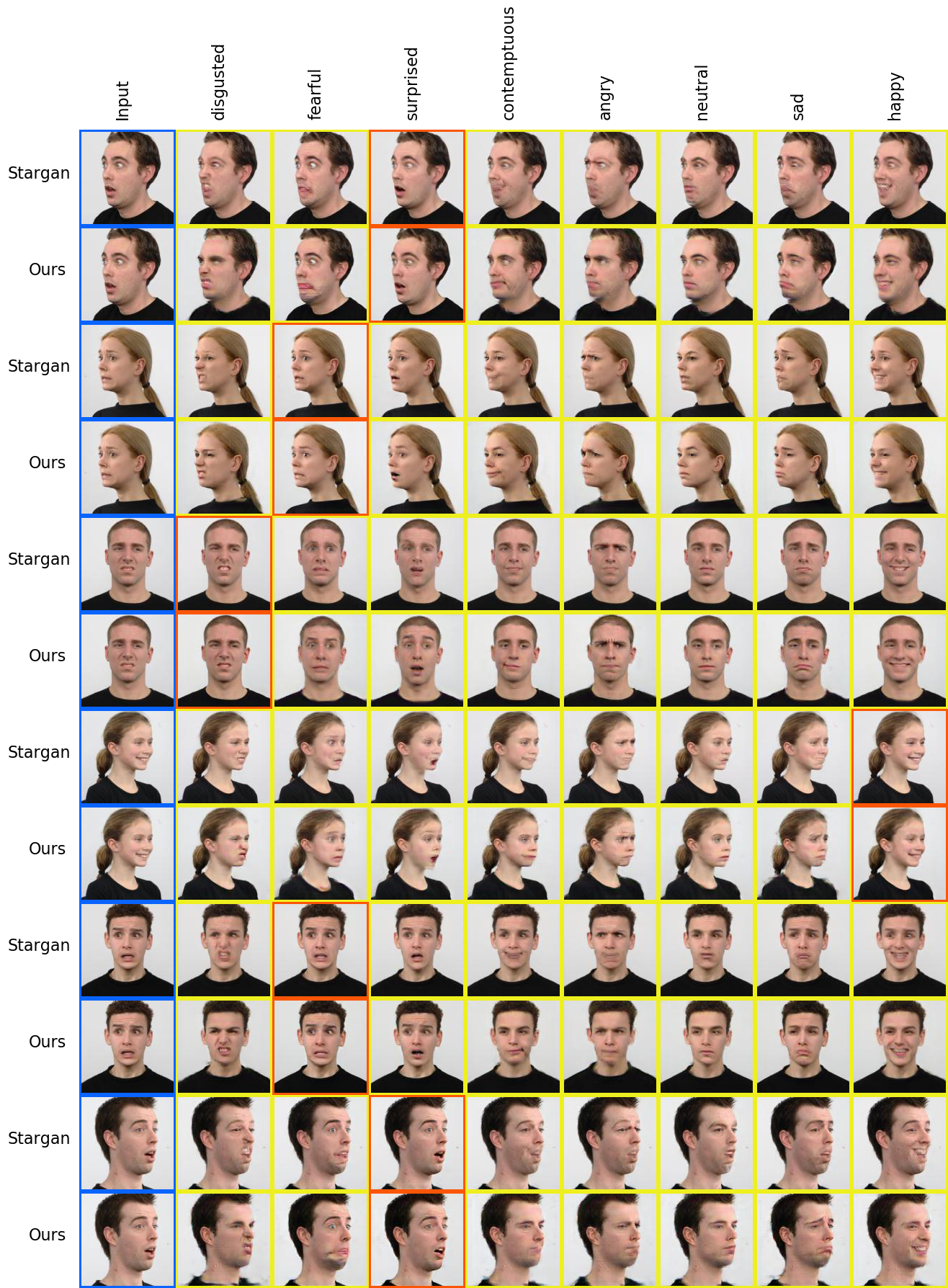}
    \caption{Qualitative comparison of attribute translation on RaFD. Blue boxed images
    are input and red boxed images are reconstructed images.}
    \label{fig:qual_rafd1}
\end{figure*}

\begin{figure*}
    \centering
    \includegraphics[trim=0cm 0cm 0cm 0cm, clip, width=1.0\textwidth]{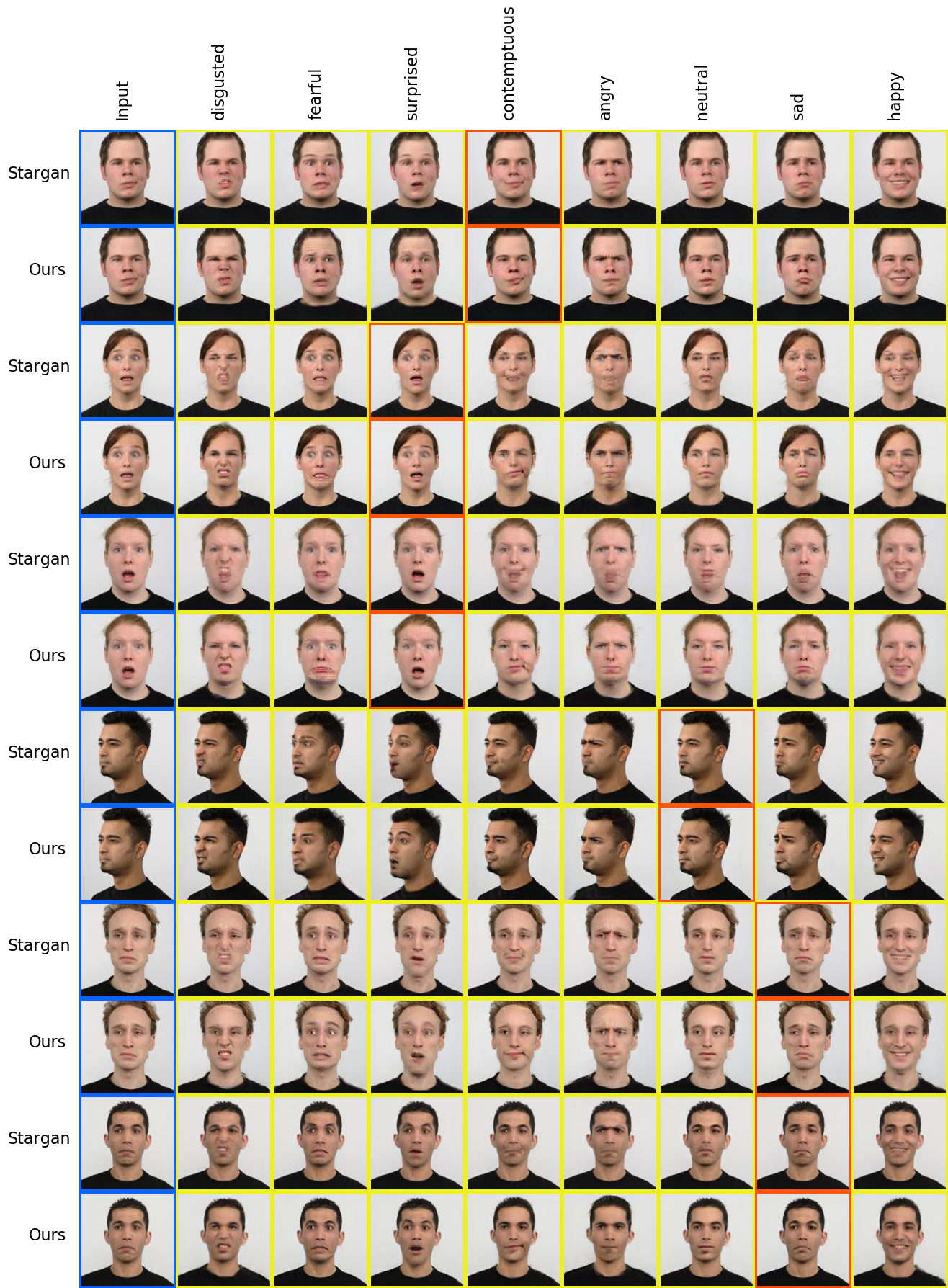}
    \caption{Qualitative comparison of attribute translation on RaFD. Blue boxed images
    are input and red boxed images are reconstructed images.}
    \label{fig:qual_rafd1}
\end{figure*}

\begin{figure*}
    \centering
    \includegraphics[trim=0cm 0cm 0cm 0cm, clip, width=1.0\textwidth]{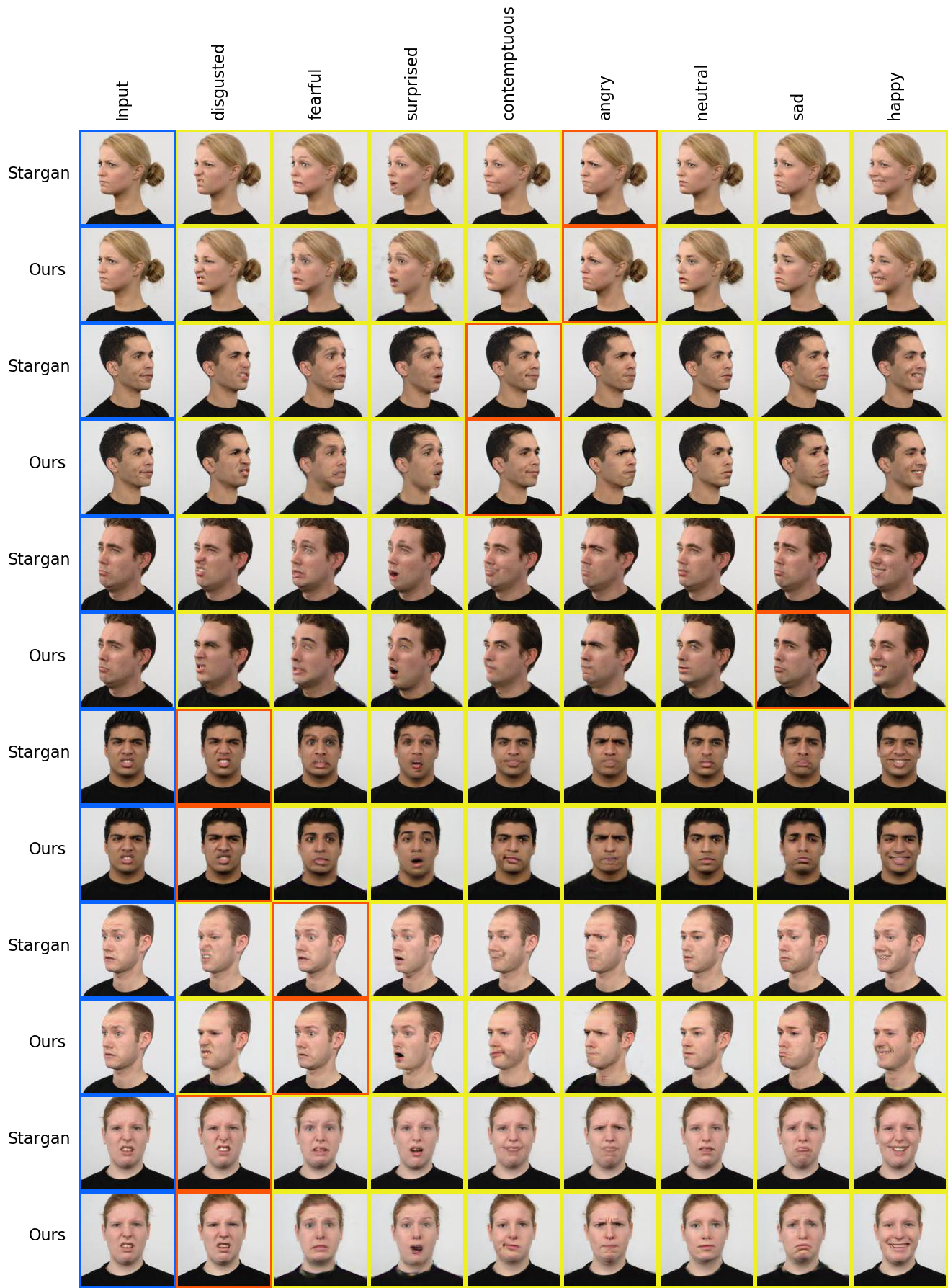}
    \caption{Qualitative comparison of attribute translation on RaFD. Blue boxed images
    are input and red boxed images are reconstructed images.}
    \label{fig:qual_rafd1}
\end{figure*}

\begin{figure*}
    \centering
    \includegraphics[trim=0cm 0cm 0cm 0cm, clip, width=1.0\textwidth]{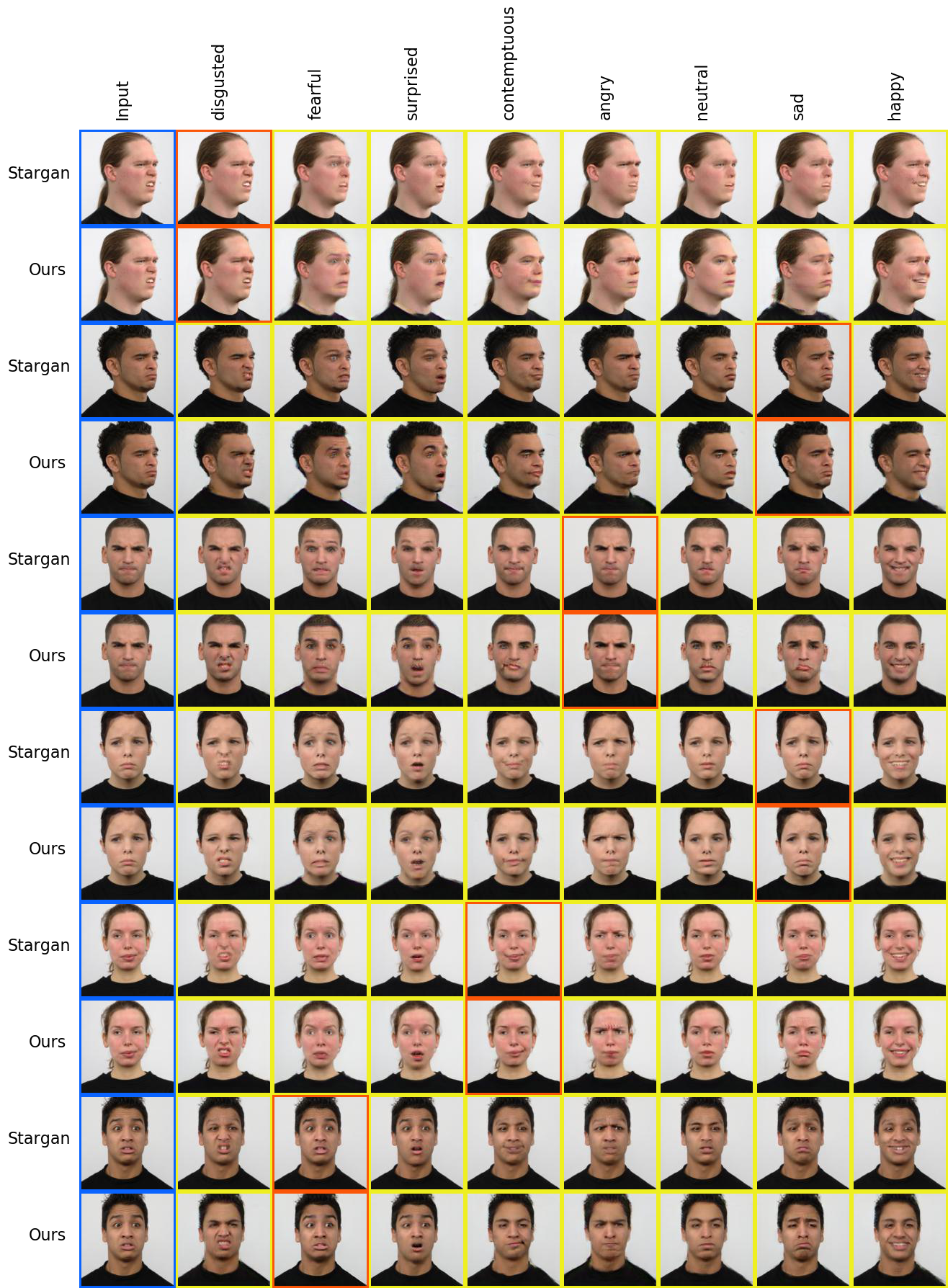}
    \caption{Qualitative comparison of attribute translation on RaFD. Blue boxed images
    are input and red boxed images are reconstructed images.}
    \label{fig:qual_rafd1}
\end{figure*}

\begin{figure*}
    \centering
    \includegraphics[trim=0cm 0cm 0cm 0cm, clip, width=1.0\textwidth]{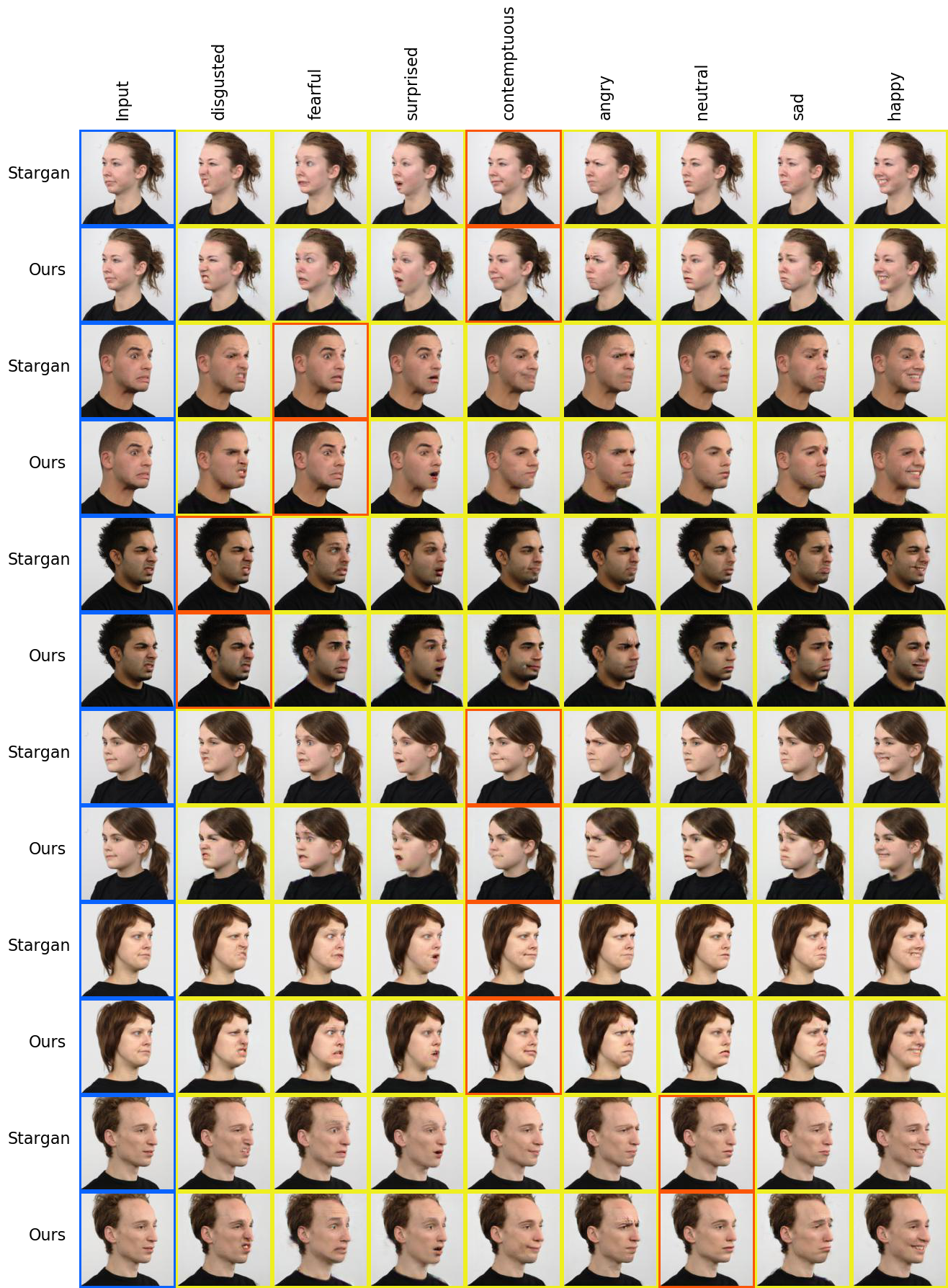}
    \caption{Qualitative comparison of attribute translation on RaFD. Blue boxed images
    are input and red boxed images are reconstructed images.}
    \label{fig:qual_rafd1}
\end{figure*}

\begin{figure*}
    \centering
    \includegraphics[trim=0cm 0cm 0cm 0cm, clip, width=1.0\textwidth]{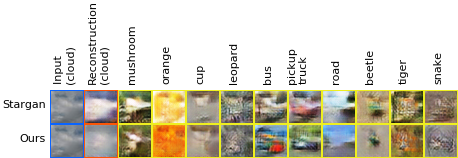}
     \includegraphics[trim=0cm 0cm 0cm 0cm, clip, width=1.0\textwidth]{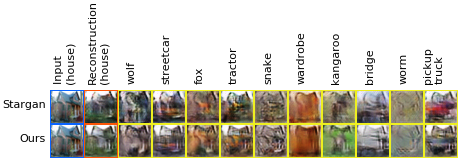}
    \includegraphics[trim=0cm 0cm 0cm 0cm, clip, width=1.0\textwidth]{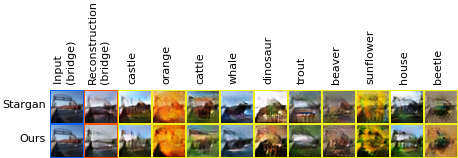}
    \includegraphics[trim=0cm 0cm 0cm 0cm, clip, width=1.0\textwidth]{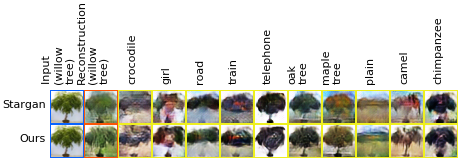}
    
    \caption{Qualitative comparison of category translation on CIFAR-100. Blue boxed images
    are input and red boxed images are reconstructed images.}
    \label{fig:qual_rafd1}
\end{figure*}

\begin{figure*}
    \centering
    \includegraphics[trim=0cm 0cm 0cm 0cm, clip, width=1.0\textwidth]{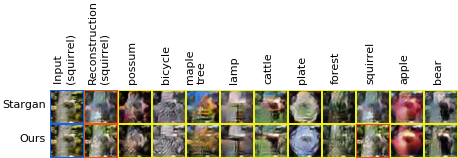}
     \includegraphics[trim=0cm 0cm 0cm 0cm, clip, width=1.0\textwidth]{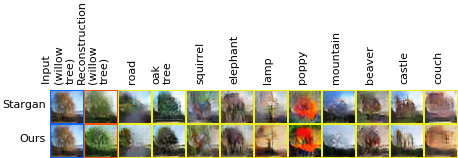}
    \includegraphics[trim=0cm 0cm 0cm 0cm, clip, width=1.0\textwidth]{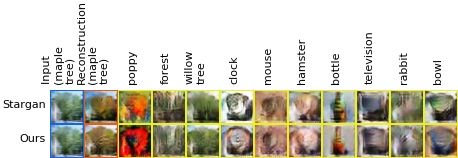}
    \includegraphics[trim=0cm 0cm 0cm 0cm, clip, width=1.0\textwidth]{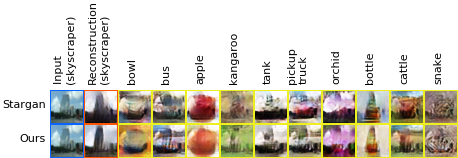}
    
    \caption{Qualitative comparison of category translation on CIFAR-100. Blue boxed images
    are input and red boxed images are reconstructed images.}
    \label{fig:qual_rafd1}
\end{figure*}

\begin{figure*}
    \centering
    \includegraphics[trim=0cm 0cm 0cm 0cm, clip, width=1.0\textwidth]{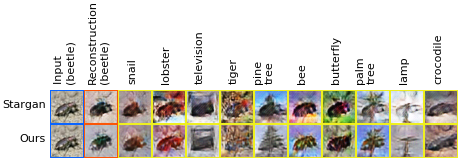}
     \includegraphics[trim=0cm 0cm 0cm 0cm, clip, width=1.0\textwidth]{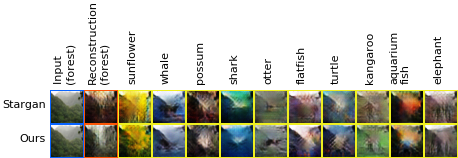}
    \includegraphics[trim=0cm 0cm 0cm 0cm, clip, width=1.0\textwidth]{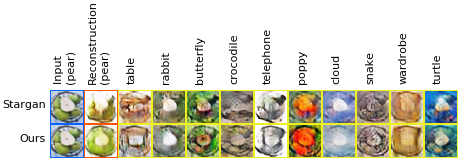}
    \includegraphics[trim=0cm 0cm 0cm 0cm, clip, width=1.0\textwidth]{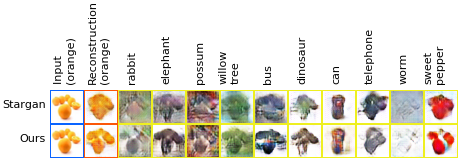}
    
    \caption{Qualitative comparison of category translation on CIFAR-100. Blue boxed images
    are input and red boxed images are reconstructed images.}
    \label{fig:qual_rafd1}
\end{figure*}
